 \documentclass[pmlr,twocolumn]{jmlr} 

\usepackage{booktabs}
\usepackage[load-configurations=version-1]{siunitx} 

\theorembodyfont{\upshape}
\theoremheaderfont{\scshape}
\theorempostheader{:}
\theoremsep{\newline}

\jmlrvolume{LEAVE UNSET}
\jmlryear{2020}
\jmlrsubmitted{LEAVE UNSET}
\jmlrpublished{LEAVE UNSET}
\jmlrworkshop{Machine Learning for Health (ML4H) 2020} 

\title[Efficient Colon Cancer Grading with GNN's]{ Efficient Colon Cancer Grading with Graph Neural Networks
}

\author{%
\Name{Franziska Lippoldt} \Email{franziska.lippoldt@augustusai.com}\\
\addr Augustus Intelligence, New York, USA
\AND
}

\begin{document}

\maketitle

\begin{abstract}
Dealing with the application of grading colorectal cancer images, this work proposes a 3 step pipeline for prediction of cancer levels from a histopathology image.
The overall model performs better compared to other state of the art methods on the colorectal cancer grading data set and shows excellent performance for the extended colorectal cancer grading set.
The performance improvements can be attributed to two main factors: The feature selection and graph augmentation method described here are spatially aware, but overall pixel position independent. Further, the graph size in terms of nodes becomes stable with respect to the model's prediction and accuracy for sufficiently large models.
The graph neural network itself consists of three convolutional blocks and linear layers, which is a rather simple design compared to other networks for this application.
\end{abstract}

This is a draft 
\section{Introduction}
The American Cancer Society’s estimate for the number of new colon cancer cases in 2020 alone is 147,950, \cite{siegel2020colorectal}. Among those, African Americans have the highest colorectal cancer incidence and mortality rates of all racial groups in the US, \cite{wolf2018colorectal}. While the reason for this are not fully understood, research on graph neural networks for colon cancer prediction based on histopathology images has the potential to facilitate better diagnosis , possibly decreased costs for a check up and better prognosis for the patient due to earlier detection. \\
The histopathology images used to identify cancer grades are microscope images of extracted tissue from a colonscopy. As the colon wall where the tissue is extracted from contains several layers, each sample might be from different layers showing different structures, possibly also different layers in one sample: The inner most layer, the so called Epitheleum, contains glands that lead to a structure that might appear similar to bubbles under the microscope, the lower layers are constituted of muscle and connection layers of different strength.  \\
While convolutional neural networks are strong in segmentation capabilities of histopathology images, this work proposes to use a cell graph to represent the image and analyse the cancer level through a graph neural network, simply because the structure of the graph and the evaluation through the network can generalize inspection of colon tissue through all of the above mentioned layers in a natural way. For general CNNs, there is typically a strong reliance on correct image size, image alignment and image scaling. Graph representations and networks have the potential to operate without pixel coordinate and those implicated dependencies.
\section{Related Work}
Related to this work in terms of \textbf{medical application} and general colon cancer prediction, a variety of convolutional neural network approaches have been proposed. Recent state of the art for colon cancer grading with convolutional neural networks has been published by \cite{shaban2020context}. The neural network described consists of a feature module with a CNN as its core that extracts features from the image into a feature cube, which is then further processed by an attention and context block to evaluate the overall grade. \\
Close to this work, \cite{zhou2019cgc} proposed a graph neural network that has a three layered approach of their adaptive GraphSage model, which contains a LSTM structure for information aggregation over three different graph convolutional layers.\\
Graph neural networks have also been used for other medical applications, publications from \cite{yu2019st} and \cite{juarez2019joint} propose to use graph neural network for 2D and or 3D image analysis of medical data, mostly scans.\\
Related to this work in terms of \textbf{graph neural network theory and architectures}, various authors have provided detailed benchmarks on aggregation methods and graph neural network architectures tested over common graph data sets such as biomedical graphs or social graphs, see \cite{dwivedi2020benchmarking}, \cite{morris2019weisfeiler} and \cite{xu2018powerful}. 
From the overall availability of different methods and architectures with performance slightly different from other graph neural networks depending on the data sets, it becomes clear that there is no one graph neural network that fits all applications. From the different aggregation methods tested on graph neural networks, this work adapts the GraphSage \cite{hamilton2017inductive} methodology for several reasons: The design to work on specifically large graphs by approximating aggregations, the capability to work in a semi supervised setting, and the large scale adaption and testing of GraphSage among other authors.\\
Prior to this, the author has been working on \cite{baylearn}, which experiments on edge downsampling methods on the MNIST Superpixel data set using sparse graphs with the intention of applying it to histopathological data. While the down sampling is also of importance to this work, it is used on nodes rather than edges, for dense graph representations rather than sparse, and hence requires a different approach to a solution.\\
Compared to the CGC Net by \cite{zhou2019cgc}, there are three main differences in this work's approach:  this paper provides neural network structures that are more simple and light weight, hence more easily reproducible in research: the feature segmentation has been trained on a u-net which does not require any backbone and can be trained in a short amount of tine. The graph neural network structure does not contain the adaptive LSTM structure but is rather a simple combination of three convolutional blocks with stabilization and pooling layers. The graph augmentation from cell features is independent of spatial coordinates, locally dense and effective in down sampling to not more than 5000 nodes per image.
\section{Cancer grading  prediction}
\begin{figure*}
\includegraphics[scale=.6]{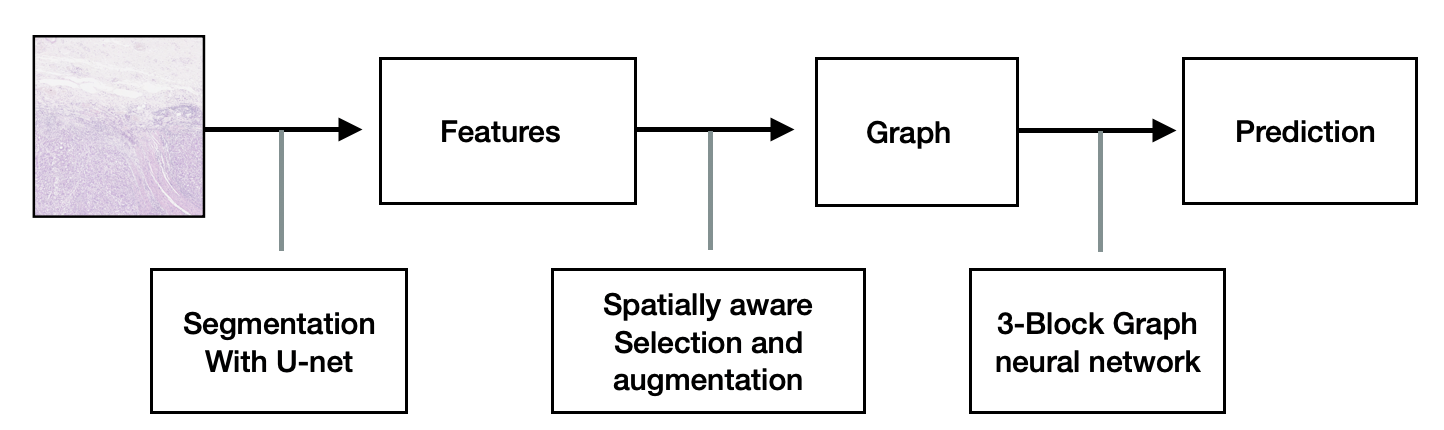}
\caption{Overall carcinoma level prediction pipeline divided into three steps}
\label{fig:pipeline}
\end{figure*}
\subsection{From medical analysis to technical application}
Cancer grading from a medical perspective requires thorough analysis of the cells within a whole tissue, and the challenge lies in interpreting a combination of the cell tissue type, tissue cut direction, cell overall distribution and abnormal structures, in addition complicated by non uniformity in the histopathology process. \\
Some of the observatory skills required to understand the complete image as one context and to differentiate between different grades are acquired by assistants through medical training, yet still hard for a technical application to understand. \\
Those challenges are amongst others:  The understanding of "empty spaces", the implicit layer type determination and the overall grading logic for an image that contains partially different cancer level.
\subsubsection{Interpretation of space and pattern} "Blank space" - in the image visible as "white", can either stand for missing tissue due to the extraction method or be a marker for a necrosis. While with the human eye there is a difference between a clean cut and a tissue decaying at the border of the cut, this feature is hard to understand from a machine learning point of view. \\
Further more glandular structures in the images pose a challenge to traditional visual analysis - \cite{awan2017glandular} have proposed to separately recognize glandular structures for cancer grading. This work assumes the opposite, namely that the graph representation in combination with the graph neural network can treat different structures for one cancer grade equally. 
This is enforced by design, as the graph representing the cell structures is locally dense, i.e. each cell is connected to each other cell, and as edge weights are applied in a locally constant manner, this means that the representation of cells around glandular structures can be topologically equivalent to a graph representing a muscular structure, if the number of cells per area is the same. For technical details see section \ref{sec:featselect} and \ref{sec:graphaug}.
\subsubsection{Comparison to a ground truth pattern}
Analysing the tissue layer type and the tissue cut direction, is crucial to differentiate between a normal and abnormal tissue. Cancer grading is not straight forward linearly related to the number of cells per image. Traditional convolutional neural networks work on a pixel wise evaluation of an image. The pipeline steps two and three however only use relational position, not absolute. Whether or not cell A is 5 more pixel to the left or the right should in no manner influence the network. This is only possible by building a graph structure based upon nodes with visual features only and passing the graph through a graph neural network.
\subsubsection{Local vs Global evaluation}
As a result of the different tissue types and layers per sample, the local cancer grading results for different regions are not in general identical, i.e. one sample might contain different grades at different areas. Which is why the graph neural network uses patching to evaluate the result of each patch locally and then reason on the overall grading system with a linear layer block, see section \ref{sec:patching}.
\subsection{Pipeline for prediction}
\begin{figure*}[t]
\includegraphics[scale=.6]{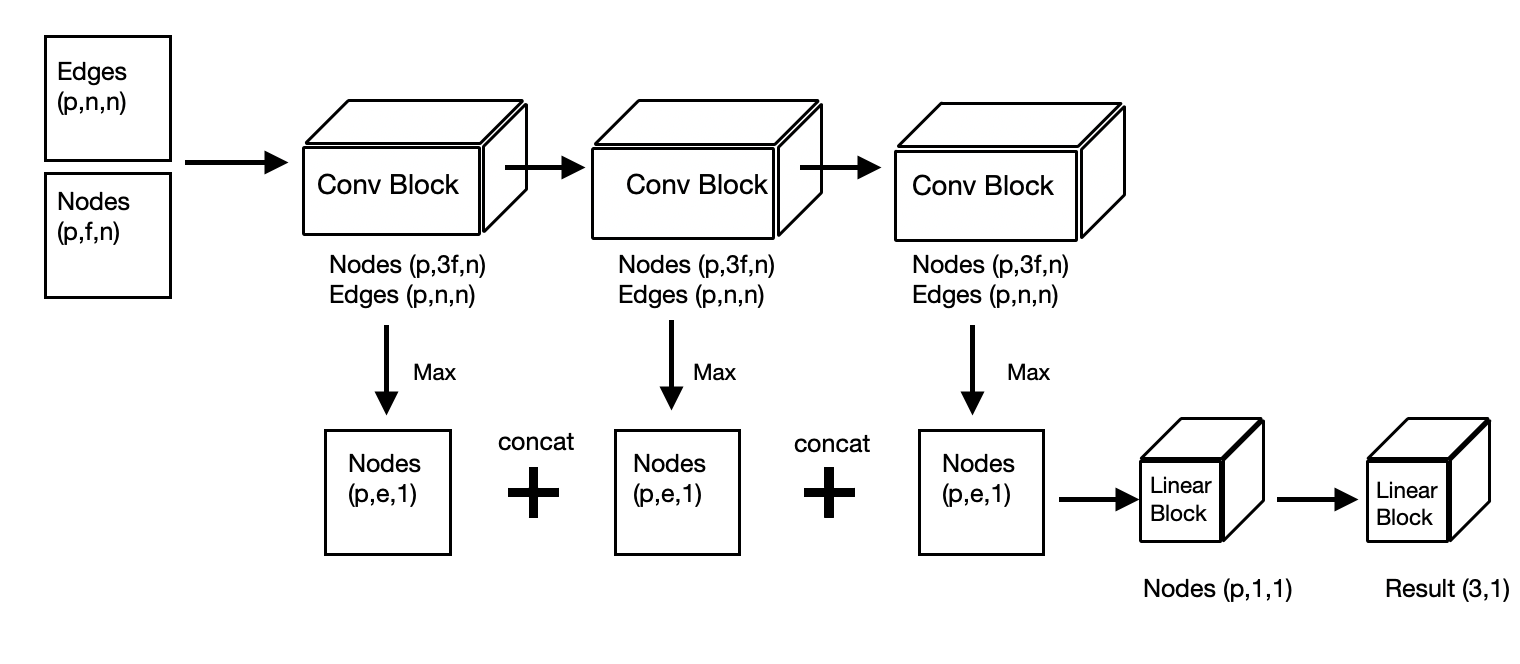}
\caption{Architecture of graph network with dimensions of vertices where as p stands for the number of patches, f for the feature dimension, e for the embedded feature dimension and n for the number of nodes.}
\label{fig:arch}
\end{figure*}

The complete pipeline from histopathology image to prediction consists of three steps, it starts with cell segmentation, which is then followed by cell graph augmentation, and concluded with level prediction through a graph neural network, see figure \ref{fig:pipeline} for reference.\\
\textbf{Step 1: Feature calculation}
This work uses a 5-layer U-net trained on cell segmentation images as originally proposed by \cite{ronneberger2015u}.
Nodes for the graph correspond to features of the cell segmentation provided by the Unet. Similar to \cite{zhou2019cgc}, 16 different properties per segmented cell are used for each node.\\
\textbf{Step 2A: Feature Selection}
Due to the high amount of cells per image, the selected number of cells are downsampled using distribution aware methods. For a fixed number of features, the selection uses random choices per area.\\
\textbf{Step 2B: Graph Augmentation}
Those randomly downsampled features are then augmented into a graph. The graph's node have 16 features each and the edges are determined using distribution based edge values. The edge value function is proportional to both the local distribution and the variance in distribution. \\
\textbf{Step 3: Graph Neural Network}
The graph is analysed through a network consisting of three convolutional blocks and a prediction layer, where as each block consists of one GraphSAGE convolution, one pooling and one normalization layer, as shown in detail in figure \ref{fig:arch}.

\section{Methodology}
The pipeline is designed to be easily reproducible and lightweight during training: the cell segmentation is achieved with a conventional U-net \cite{ronneberger2015u}, the feature selection and graph augmentation methods have linear complexity with respect to the number of nodes (i.e. cells), the graph neural network acts on vertices and edges in 3 convolutional blocks. In the following, the different technical details of those methods will be described:
\subsection{Cell feature calculation}
\label{sec:cellfeat}
A U-net with 5 convolutional layers is used as described by the authors in \cite{ronneberger2015u}. The resulting mask is processed to extract 16 features per cell instance, as proposed in \cite{zhou2019cgc}. The first half of those are bounding box color features, the second half are segmentation mask based image features. In the experiments a study on the feature dimension has been executed.
\subsection{Spatially aware feature selection}
\label{sec:featselect}
For a given segmentation map $s(i,j)$ of an image $I$ of dimension $w \times h $ at pixel position $(i,j)$ , let us define the distribution map as $D(i,j)$, which uses a regular 2D grid of dimension $dxd$ and counts the number of points per box.
Then for any set of feature points $ F = \{ f_0,\dots,f_N \}$ the corresponding distribution map for M feature points is 
\begin{equation*}
D_M(i,j) = \frac{M}{\| D \|} D(i,j)
\end{equation*}
where $ \| \cdot \|$ stands for the $L_1$ matrix norm.
Given the number of points per area through the distribution map $D_M(i,j)$, the features are augmented into a graph through random selection of nodes per area.

\subsection{Spatially aware graph augmentation}
\label{sec:graphaug}
For a given set of feature points  $ F = \{ f_0,\dots,f_N \}$ , the edge value between two feature points $f_k,f_m$ and their corresponding coordinates $c_k,c_m$ is defined as:
\begin{equation*}
\begin{split}
A_{k,m} &= \alpha (D_M(c_k)+D_M(c_m)) \\
        &+  \beta  | D_M(c_k) - D_M(c_m) |
\end{split}
\end{equation*}
where in this case  $|\cdot |$ stands for the absolute value and $\alpha,\beta \in [0,1]$ are hyper parameters. 
Then the Matrix $A$ corresponds to the adjacency matrix of the augmented graph.

\subsection{Cell Graph Neural Network architecture}
This part of the pipeline leverages a graph neural network architecture that acts both on nodes and edges, applies pooling and normalization operations and classifies the graph into a category. \\
It consists of two parts, the graph modification and the graph classification. The graph modification network consists of graph convolution blocks that modify both edge and vertices stepwise. Each convolutional block consists of three graph convolutions that are concatenated, one pooling and one normalizing layer.\\
As a graph convolution propagation rule, this work uses GraphSAGE \cite{hamilton2017inductive}: 
\begin{equation*}
h_v^k \leftarrow \sigma ( \textbf{W}  \cdot \textit{mean} ( \{ h_v^{k-1} \} \cup \{ h_u^{k-1} , \forall \textit{u} \in \textsl{N} \}  ) )  
\end{equation*}
where W is the edge weight matrix, N stands for the set of points lying in the direct neighborhood of node v, and h denotes the node's feature with respect to node v and number of neighborhood jumps k. In this work only one hop convolutions will be applied.
\subsubsection{Feature embedding}
Many graph neural network architectures do not work with the original nodes i.e. features as input but upscale the dimensions, such as shown in work by \cite{errica2019fair}.
Upscaling of the feature dimensions is done through increasing the output dimensions of the graph convolution respectively. Down scaling of features is executed by applying linear layers with the output dimension as the desired scaled size. \\
While each convolutional block contains three convolutions that are being concatenated, this would increase the feature dimensions by a factor equal to the number of concatenations. The feature dimension of each convolutional block is hence down scaled back to the original embedding dimension with a linear layer at each step.
The convolutional block structure is simplified compared to the work proposed in \cite{zhou2019cgc}, as the bi-directional LSTM is not used this way.
\subsubsection{GNN normalization and stabilization}
Graph neural networks suffer from the over smoothening of adjacency values. To fix this, the values of the adjacency matrix $A$ are reassigned after every block in the following manner as originally proposed by \cite{chen2019multi}:\\
\begin{equation}
A'_{i,j} = 
\begin{cases} 
1-p \textbf{ if } i = j\\ 
\frac{p}{\sum_{i, i\neq j }A_{i,j}} \textbf{ if } i \neq  j\\ 
\end{cases}
\end{equation}
\subsubsection{GNN pooling operations}
The graph neural network leverages differential graph pooling Diffpool as originally described in \cite{ying2018hierarchical}, which computes an assignment matrix seperating nodes into different clusters.
\subsubsection{Graph patching}
\label{sec:patching}
Given the original Graph G corresponding to the image I, the nodes and respective edges are split into patches according to the pixel position along each axis into two sections resulting in four patches. The patches are each treated as individual graphs during the convolutional blocks and are merged into one result at the end of the pipeline with a block of linear layers.\\
For sake of simplicity in our architecture sketch, it is assumed that the patches have same size or are padded with zeros if not. In practice different number of nodes per patch have been used, as the splitting method does not take number of nodes into account.\\
Instead of typically used batch norm, the graph convolutional layers' outputs are stabilized with "patch norm".

\section{Experiments}
\subsection{Preliminaries}
All of the experiments are conducted on a single GPU per task. Many of the experiments take less than 20 GB of GPU Memory, some experiments on the single graph with high number of nodes take up to 46 GB. All experiments on work from this paper use less than  10 GB of CPU memory. Graph augmentation was implemented to work on multiple cores.\\
The neural network has been coded in PyTorch \cite{paszke2019pytorch} and specifically PyTorch Geometric \cite{fey2019fast}. For the cell segmentation the code available at \cite{unetgh} has been used. 

\subsection{Data}
The Unet for semantic segmentation of cells is trained on the training set of the \textbf{colorectal cancer cell data set} CoNSeP \cite{graham2019hover}, which contains 41 images with segmentation labels of resolution 1000x1000 pixel.\\
 The graph neural network is trained on the \textbf{Extended Colorectal Adenocarcinoma Dataset} (abbreviated in this work with ECA) for Grading of Histology Images \cite{shaban2020context}, which classifies 300 images from 178 different patients of resolutions around 5000x7000 pixel into three categories: no cancer , low level and high level.
 In order to enable comparisons to scores in previous works, the \textbf{Colorectal Cancer Grading Dataset} (CRC) ( \cite{awan2017glandular}) is used, which contains a subset of the Extended Colorectal Adenocarcinoma Dataset, in total 137 images from 38 different patients.

\subsection{Feature calculation} 
For the semantic segmentation, the u-net is trained on 100 epochs on the CoNSeP training data set. The training and test split is provided with the data. The binary cross entropy with logits is used for training, together with a learning rate plateau scheduler and Root Mean Square Propagation. After those epochs, it overall achieves a dice test score of 75.60 out of 100. As the dice score heavily punishes segmentation areas larger than the labeled segmentation, this means that the segmented cells potentially are slightly larger than the ground truth. Cells in the training data and test data have a bounding of around 10x10 pixel to 20x20 pixel, with main variations in different images and small size variation only per one image. For a 10x10 sized bounding box, with a mask area smaller than 100 pixel, a dice score of around 75 corresponds to a cell of around 125 pixel area. 
\begin{table}
  \caption{Overview over the ECA (\cite{shaban2020context}) and CRC datasets (\cite{awan2017glandular}) used for cancer level prediction; the label split shows the distribution of samples among labels no grade, low grade, high grade and the fold split shows the number of samples per fold 1,2 and 3}
  \label{table:data}
  \centering
  \begin{tabular}{l|l|l}
    \toprule
   Dataset & CRC & ECA  \\
    
    \midrule
   Samples & 137 & 300  \\
   Label Split & 71-31-35 & 120-120-60 \\
   Fold Split & 45-46-46 & 100-100-100  \\
   Patients & 38 & 178  \\

    \bottomrule
  \end{tabular}
\end{table}
\subsection{Cancer Grading and Graph Network}

\begin{table*}[ht]
  \caption{Overall best accuracy scores for different parameters, Accuracy and standard deviation values are provided in percentage}
  \label{table:acctotal}
  \centering
  \begin{tabular}{lllll}
    \toprule
   Method &    CRC & ECA  \\
    \midrule
    CA-CNN \cite{shaban2020context} & $95.70 \pm 3.04$ & - \\
    CGC \cite{zhou2019cgc} & $ 97.00 \pm 1.10$ & -  \\
    \midrule 
    this work (CRC) & $99.26 \pm 1.05$ & - \\
    this work (ECA)& $99.26 \pm 0.66$ & $99.33 \pm 0.94$ \\
    \bottomrule
  \end{tabular}
\end{table*}

In Table \ref{table:acctotal}, it is shown that this work improves the accuracy of other works on this data set by around two percent on the CRC data set and the ECA data set.
\textbf{Parameter}
The normalization layer is being used with p = 0.4. The embedding dimension has been chosen to be 100, in some experiments 120. From the classification head, the first linear layer block contains three linear layers with dimensions 50,25,3 respectively. The patch classification layer contains two linear layers with dimensions 3 and 1. The learning rate has to be adjusted for each specific experiment but on average lies between 0.00005 and 0.00001 at initialization. The learning rate is scheduled with a plateau scheduler with respect to the accuracy\\
\textbf{Evaluation criteria}
As training is executed using 3-fold cross validation, the evaluation for the complete pipeline is given in terms of the average and standard deviation of the last three accuracy values. The loss function used during training of the graph neural network is a smooth L1 loss.\\
\textbf{Time estimations}
Training on the graph neural network including all necessary steps requires around 6-14h per model and configuration, with variations of several hours for  usage of normalization layers and large graph sizes.
\subsection{Ablation Studies}
\subsubsection{Patching versus one graph}

\begin{figure}
\includegraphics[scale=.5]{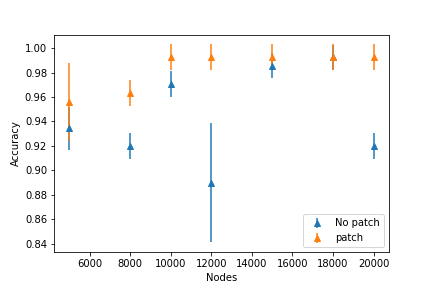}
\caption{Comparison of accuracy scores with and without patching for different number of nodes on the CRC dataset}
\label{fig:patchwhole}
\end{figure}
In figure \ref{fig:patchwhole}, accuracy values for one graph per image vs. 4 graphs per image have been compared.\\
While one graph per image yields a straight forward classification score, separating the graph into patches and evaluating on those patches requires a classification head that decides on the overall label based on each patch. Evaluation results for a single graph are unstable over the selection of number of nodes for the augmentation and achieve in general lower accuracy than with four graphs per image.
\subsubsection{Different cell segmentation features as input}
\begin{figure}
\includegraphics[scale=.5]{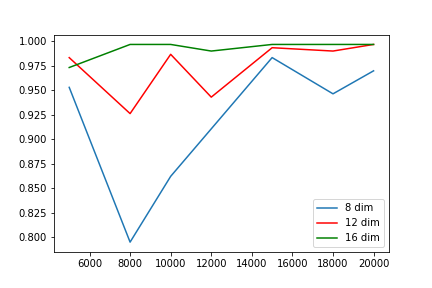}
\caption{Change of accuracy for different feature dimensions 8,12,16 of nodes and different graph sizes for training on the ECA dataset}
\label{fig:featdim}
\end{figure}
While this paper adapted the 16 cell segmentation features as in \cite{zhou2019cgc}, a set of experiments have been conducted on using smaller amount of cell features from the segmentation. Of those features, the first half of them characterize the bounding box color and black and white values, while the second half of features characterizes the properties of the masked cell. Testing has been done on 8-dim feat input (only looking at bounding box based features) and 12 dim input with restricted segmentation features. While the stability inspected over smaller graphs decreases for lower feature dimensions, for a graph size of 16000 the accuracy values are observed to peak for all feature dimensions and differ by around 2 percent only. 

\section{Conclusion}
In this work a stable and efficient graph neural network pipeline has been proposed to evaluate cancer grades. Experiments on parameters have shown that a stable architecture design and efficient pipeline can improve evaluation results.\\
Graph augmentation, just like image augmentation and text augmentation, has significant impact on the quality of the results. Compared to the state of the art, the pipeline accuracy results are improved by around two percent and the overall number of false predictions is decreased. \\
Further, experiments have been executed with respect to graph size and number of nodes, which shows that downsampling of graphs must not lead to a loss in accuracy. This would also possibly enable testing on devices and hardware with less available memory, as graph neural network training can be much more expensive than training of traditional neural networks. \\
Further, the author suggest that this pipeline has the potential to be used on other diseases related to histopathological image diagnosis.
\section{Acknowledgement}
The author would like to thank Dr. Anish Mohammed for this medical advice. Special thanks to Gregory Senay for his support during the writing process.

\newpage
\bibliography{paper.bib}

\begin{thebibliography}{20}
\providecommand{\natexlab}[1]{#1}
\providecommand{\url}[1]{\texttt{#1}}
\expandafter\ifx\csname urlstyle\endcsname\relax
  \providecommand{\doi}[1]{doi: #1}\else
  \providecommand{\doi}{doi: \begingroup \urlstyle{rm}\Url}\fi

\bibitem[Awan et~al.(2017)Awan, Sirinukunwattana, Epstein, Jefferyes, Qidwai,
  Aftab, Mujeeb, Snead, and Rajpoot]{awan2017glandular}
Ruqayya Awan, Korsuk Sirinukunwattana, David Epstein, Samuel Jefferyes, Uvais
  Qidwai, Zia Aftab, Imaad Mujeeb, David Snead, and Nasir Rajpoot.
\newblock Glandular morphometrics for objective grading of colorectal
  adenocarcinoma histology images.
\newblock \emph{Scientific reports}, 7\penalty0 (1):\penalty0 1--12, 2017.

\bibitem[Chen et~al.(2019)Chen, Wei, Wang, and Guo]{chen2019multi}
Zhao-Min Chen, Xiu-Shen Wei, Peng Wang, and Yanwen Guo.
\newblock Multi-label image recognition with graph convolutional networks.
\newblock In \emph{Proceedings of the IEEE Conference on Computer Vision and
  Pattern Recognition}, pages 5177--5186, 2019.

\bibitem[Dwivedi et~al.(2020)Dwivedi, Joshi, Laurent, Bengio, and
  Bresson]{dwivedi2020benchmarking}
Vijay~Prakash Dwivedi, Chaitanya~K Joshi, Thomas Laurent, Yoshua Bengio, and
  Xavier Bresson.
\newblock Benchmarking graph neural networks.
\newblock \emph{arXiv preprint arXiv:2003.00982}, 2020.

\bibitem[Errica et~al.(2019)Errica, Podda, Bacciu, and Micheli]{errica2019fair}
Federico Errica, Marco Podda, Davide Bacciu, and Alessio Micheli.
\newblock A fair comparison of graph neural networks for graph classification.
\newblock \emph{arXiv preprint arXiv:1912.09893}, 2019.

\bibitem[Fey and Lenssen(2019)]{fey2019fast}
Matthias Fey and Jan~Eric Lenssen.
\newblock Fast graph representation learning with pytorch geometric.
\newblock \emph{arXiv preprint arXiv:1903.02428}, 2019.

\bibitem[Graham et~al.(2019)Graham, Vu, Raza, Azam, Tsang, Kwak, and
  Rajpoot]{graham2019hover}
Simon Graham, Quoc~Dang Vu, Shan E~Ahmed Raza, Ayesha Azam, Yee~Wah Tsang,
  Jin~Tae Kwak, and Nasir Rajpoot.
\newblock Hover-net: Simultaneous segmentation and classification of nuclei in
  multi-tissue histology images.
\newblock \emph{Medical Image Analysis}, 58:\penalty0 101563, 2019.

\bibitem[Hamilton et~al.(2017)Hamilton, Ying, and
  Leskovec]{hamilton2017inductive}
Will Hamilton, Zhitao Ying, and Jure Leskovec.
\newblock Inductive representation learning on large graphs.
\newblock In \emph{Advances in neural information processing systems}, pages
  1024--1034, 2017.

\bibitem[Juarez et~al.(2019)Juarez, Selvan, Saghir, and
  de~Bruijne]{juarez2019joint}
Antonio Garcia-Uceda Juarez, Raghavendra Selvan, Zaigham Saghir, and Marleen
  de~Bruijne.
\newblock A joint 3d unet-graph neural network-based method for airway
  segmentation from chest cts.
\newblock In \emph{International Workshop on Machine Learning in Medical
  Imaging}, pages 583--591. Springer, 2019.

\bibitem[Lippoldt and Lavin(2020)]{baylearn}
Franziska Lippoldt and Alexander Lavin.
\newblock Attention-sampling graph convolutional networks.
\newblock submitted to Arxiv, 2020.

\bibitem[Milesi(2020)]{unetgh}
Alexandre Milesi.
\newblock Unet: semantic segmentation with pytorch, 2020.
\newblock URL \url{https://github.com/milesial/Pytorch-UNet}.

\bibitem[Morris et~al.(2019)Morris, Ritzert, Fey, Hamilton, Lenssen, Rattan,
  and Grohe]{morris2019weisfeiler}
Christopher Morris, Martin Ritzert, Matthias Fey, William~L Hamilton, Jan~Eric
  Lenssen, Gaurav Rattan, and Martin Grohe.
\newblock Weisfeiler and leman go neural: Higher-order graph neural networks.
\newblock In \emph{Proceedings of the AAAI Conference on Artificial
  Intelligence}, volume~33, pages 4602--4609, 2019.

\bibitem[Paszke et~al.(2019)Paszke, Gross, Massa, Lerer, Bradbury, Chanan,
  Killeen, Lin, Gimelshein, Antiga, et~al.]{paszke2019pytorch}
Adam Paszke, Sam Gross, Francisco Massa, Adam Lerer, James Bradbury, Gregory
  Chanan, Trevor Killeen, Zeming Lin, Natalia Gimelshein, Luca Antiga, et~al.
\newblock Pytorch: An imperative style, high-performance deep learning library.
\newblock In \emph{Advances in neural information processing systems}, pages
  8026--8037, 2019.

\bibitem[Ronneberger et~al.(2015)Ronneberger, Fischer, and
  Brox]{ronneberger2015u}
Olaf Ronneberger, Philipp Fischer, and Thomas Brox.
\newblock U-net: Convolutional networks for biomedical image segmentation.
\newblock In \emph{International Conference on Medical image computing and
  computer-assisted intervention}, pages 234--241. Springer, 2015.

\bibitem[Shaban et~al.(2020)Shaban, Awan, Fraz, Azam, Tsang, Snead, and
  Rajpoot]{shaban2020context}
Muhammad Shaban, Ruqayya Awan, Muhammad~Moazam Fraz, Ayesha Azam, Yee-Wah
  Tsang, David Snead, and Nasir~M Rajpoot.
\newblock Context-aware convolutional neural network for grading of colorectal
  cancer histology images.
\newblock \emph{IEEE Transactions on Medical Imaging}, 2020.

\bibitem[Siegel et~al.(2020)Siegel, Miller, Goding~Sauer, Fedewa, Butterly,
  Anderson, Cercek, Smith, and Jemal]{siegel2020colorectal}
Rebecca~L Siegel, Kimberly~D Miller, Ann Goding~Sauer, Stacey~A Fedewa, Lynn~F
  Butterly, Joseph~C Anderson, Andrea Cercek, Robert~A Smith, and Ahmedin
  Jemal.
\newblock Colorectal cancer statistics, 2020.
\newblock \emph{CA: a cancer journal for clinicians}, 2020.

\bibitem[Wolf et~al.(2018)Wolf, Fontham, Church, Flowers, Guerra, LaMonte,
  Etzioni, McKenna, Oeffinger, Shih, et~al.]{wolf2018colorectal}
Andrew~MD Wolf, Elizabeth~TH Fontham, Timothy~R Church, Christopher~R Flowers,
  Carmen~E Guerra, Samuel~J LaMonte, Ruth Etzioni, Matthew~T McKenna, Kevin~C
  Oeffinger, Ya-Chen~Tina Shih, et~al.
\newblock Colorectal cancer screening for average-risk adults: 2018 guideline
  update from the american cancer society.
\newblock \emph{CA: a cancer journal for clinicians}, 68\penalty0 (4):\penalty0
  250--281, 2018.

\bibitem[Xu et~al.(2018)Xu, Hu, Leskovec, and Jegelka]{xu2018powerful}
Keyulu Xu, Weihua Hu, Jure Leskovec, and Stefanie Jegelka.
\newblock How powerful are graph neural networks?
\newblock \emph{arXiv preprint arXiv:1810.00826}, 2018.

\bibitem[Ying et~al.(2018)Ying, You, Morris, Ren, Hamilton, and
  Leskovec]{ying2018hierarchical}
Zhitao Ying, Jiaxuan You, Christopher Morris, Xiang Ren, Will Hamilton, and
  Jure Leskovec.
\newblock Hierarchical graph representation learning with differentiable
  pooling.
\newblock In \emph{Advances in neural information processing systems}, pages
  4800--4810, 2018.

\bibitem[Yu et~al.(2019)Yu, Yin, and Zhu]{yu2019st}
Bing Yu, Haoteng Yin, and Zhanxing Zhu.
\newblock St-unet: A spatio-temporal u-network for graph-structured time series
  modeling.
\newblock \emph{arXiv preprint arXiv:1903.05631}, 2019.

\bibitem[Zhou et~al.(2019)Zhou, Graham, Koohbanani, Shaban, Heng, and
  Rajpoot]{zhou2019cgc}
Yanning Zhou, Simon Graham, Navid~Alemi Koohbanani, Muhammad Shaban, Pheng-Ann
  Heng, and Nasir Rajpoot.
\newblock Cgc-net: Cell graph convolutional network for grading of colorectal
  cancer histology images.
\newblock In \emph{The IEEE International Conference on Computer Vision (ICCV)
  Workshops}, 2019.

\end{thebibliography}
\end{document}